# Benchmarking Federated Learning Frameworks for Medical Imaging Deployment: A Comparative Study of NVIDIA FLARE, Flower, and Owkin Substra

Riya Gupta
Harvard T.H. Chan School of Public Health, Boston, MA, USA

Sahil Nalawade
Dana-Farber Cancer Institute, Boston, MA, USA

Alexander Chowdhury
Dana-Farber Cancer Institute, Boston, MA, USA

Correspondence: priyalgupta@hsph.harvard.edu; Sahil_Nalawade@dfci.harvard.edu

## Abstract

Federated Learning (FL) has emerged as a transformative paradigm in medical AI, enabling collaborative model training across institutions without direct data sharing. This study benchmarks three prominent FL frameworks NVIDIA FLARE, Flower, and Owkin Substra to evaluate their suitability for medical imaging applications in real-world settings. Using the PathMNIST dataset, we assess model performance, convergence efficiency, communication overhead, scalability, and developer experience. Results indicate that NVIDIA FLARE offers superior production scalability, Flower provides flexibility for prototyping and academic research, and Owkin Substra demonstrates exceptional privacy and compliance features. Each framework exhibits strengths optimized for distinct use cases, emphasizing their relevance to practical deployment in healthcare environments.

## Introduction

As medical data continues to grow in scale and complexity, Federated Learning (FL) offers a promising approach to train robust AI models while respecting privacy regulations such as HIPAA and GDPR. By enabling institutions to collaborate on model training without exposing raw data, FL bridges the gap between data utility and security, which remain critical concerns in medical AI applications.

The present work benchmarks three FL frameworks widely adopted in healthcare research and deployment: NVIDIA FLARE (NVIDIA, 2024), Flower (Flower, 2024), and Owkin Substra (Owkin, 2024). While these systems share common principles of decentralized model training, their technical implementations differ in architecture, deployment complexity, privacy features, and scalability. Through a standardized experimental setup and parallel evaluation across frameworks, this study provides a practical comparison of their strengths and limitations for medical imaging use cases, emphasizing considerations relevant to real-world deployment in healthcare settings.

## Methods

### *Experimental Setup*

All experiments were performed on Google Cloud Platform (GCP). The baseline configuration consisted of an 8-core CPU environment; subsequent tests leveraged GPU-enabled instances for computational scaling. Each FL experiment used five clients, each performing three communication rounds with two local epochs per round to ensure consistency across frameworks. Framework-specific Python versions and environment dependencies were maintained separately to avoid conflicts. The same data preprocessing, model, and training configuration were applied across all frameworks to ensure a fair comparison.

### *Dataset and Task*

The PathMNIST dataset was employed for this study (Yang et al., 2022). It is a colon pathology dataset comprising approximately 100,000 images categorized into nine classes from the MedMNIST v2 collection. All images were resized to 28×28 and normalized to (0.5, 0.5, 0.5) per channel. A CNN-based architecture was implemented for the classification task. The model consisted of two 5×5 convolutional layers with ReLU activations and 2×2 max pooling, followed by fully connected layers of 120 and 84 units, and a nine-class output layer. Training used stochastic gradient descent (learning rate = 0.001, momentum = 0.9, batch size = 128) with standard FedAvg aggregation.

*Frameworks Evaluated*

- NVIDIA FLARE: Industrial-grade FL framework emphasizing production scalability, advanced orchestration, and integration with MLflow, TensorBoard, and Weights & Biases.
- Flower: Research-oriented platform enabling lightweight simulations, fast prototyping, and flexible integration into existing ML codebases.
- Owkin Substra: Enterprise-grade system designed for strict compliance, incorporating built-in differential privacy, encryption, audit logging, and isolated executor nodes for regulated medical environments.

*Metrics for Comparison*

Evaluation metrics included:
- Model performance: Accuracy, F1 score, and AUC
- Convergence efficiency: Rounds to reach 75% accuracy
- Communication efficiency: Data transferred per round
- Resource utilization: Total computational time (CPU/GPU hours) and memory footprint
- Qualitative factors: Developer experience, setup complexity, and scalability

## Results

The results in Tables 1 and 2 summarize trade-offs between scalability, ease of use, and privacy-related capabilities.

All results reported below are from the PathMNIST classification task for a five-client setup.

*Quantitative Evaluation*

| Metric | NVIDIA FLARE | Flower | Owkin Substra |
| --- | --- | --- | --- |
| Final Accuracy (%) | ~70 | ~70 | ~60 |
| Rounds to Converge | 10 | 10 | 9 |
| Training Time (hrs) | High | Medium | Low to Medium |
| Communication Overhead | High | Low | Medium to High |
| Stated Scalability (from framework documentation) | Thousands | Hundreds to Thousands | Dozens |

*Table 1*. Quantitative evaluation of FL frameworks on PathMNIST. Note: 'Rounds to Converge' is defined as rounds to reach 75% accuracy.

*Qualitative Assessment*

| Category | NVIDIA FLARE | Flower | Owkin Substra |
|---|---|---|---|
| Setup Complexity | High | Low | Medium |
| Debugging and Logging | Easy | Medium | Hard |
| Privacy and Security Features | Moderate | Low | High |
| Deployment Readiness | High | Low | Medium |

*Table 2.* Qualitative assessment of framework usability and deployment readiness.

*Figures & Visualizations*

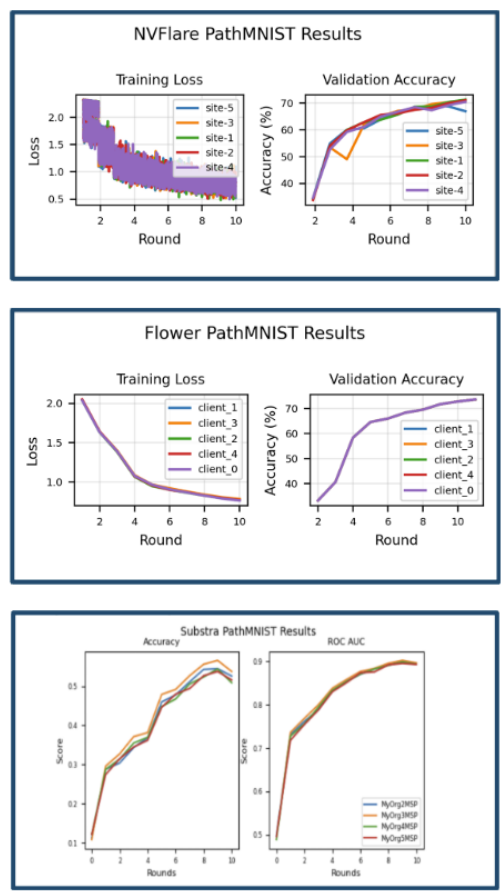

Figure 1: Comparative PathMNIST results across federated learning frameworks.

These visualizations illustrate how the three federated learning frameworks behave during training on the PathMNIST dataset. NVIDIA FLARE shows the fastest and most stable convergence, reaching high validation accuracy within fewer communication rounds but with greater computational demand. Flower demonstrates comparable final accuracy with lower communication overhead, reflecting its lightweight, research-oriented design. Substra converges more gradually yet maintains consistent and stable performance across clients, emphasizing its focus on secure and compliance-driven operation. Together, these trends complement the quantitative and qualitative findings, highlighting trade-offs between scalability, efficiency, and privacy features among the frameworks.

## Discussion

This comparative analysis demonstrates that there is no single best FL framework; rather, each excels in distinct operational contexts. NVIDIA FLARE is optimal for hospital-scale deployments and clinical workflows, given its orchestration tools, scalability, and reliability (NVIDIA, 2024). Flower is best suited for academic research and rapid experimentation, as its lightweight design and minimal setup enable fast iteration (Flower, 2024). Owkin Substra prioritizes compliance-driven privacy for regulated collaborations but requires more setup effort and offers less flexibility for integration (Owkin, 2024).

These findings emphasize that framework selection should reflect project maturity, deployment goals, and regulatory requirements. For large-scale or production implementations, NVIDIA FLARE provides the most robust infrastructure. For exploratory or research-oriented work, Flower remains the most practical choice. In contrast, Substra is most appropriate for regulated environments where privacy, auditability, and data governance are central to operational success.

## Conclusion

Federated Learning represents a cornerstone of privacy-preserving medical AI, and benchmarking frameworks such as NVIDIA FLARE, Flower, and Owkin Substra remains essential for understanding their readiness in real-world healthcare applications. This study finds that:
- NVIDIA FLARE is the most production-ready and scalable system,
- Flower is the most flexible for research and prototyping, and
- Owkin Substra is the most compliant for regulated environments.

The results provide practical guidance for selecting FL frameworks based on intended use and deployment requirements. Future work could include additional medical imaging datasets and the evaluation of privacy-preserving methods such as secure aggregation and federated differential privacy to further assess scalability across institutional settings.